**AI Needs Physics More Than Physics Needs AI**

By Peter Coveney* and Roger Highfield

**Abstract**

Artificial intelligence (AI) is commonly depicted as transformative. Yet, after more than a decade of hype, its measurable impact remains modest outside a few high-profile scientific and commercial successes. The 2024 Nobel Prizes in Chemistry and Physics recognized AI's potential, but broader assessments indicate the impact to date is often more promotional than technical. We argue that while current AI may influence physics, physics has significantly more to offer this generation of AI. Current architectures—large language models, reasoning models, and agentic AI – can depend on trillions of meaningless parameters, suffer from distributional bias, lack uncertainty quantification, provide no mechanistic insights, and fail to capture even elementary scientific laws. We review critiques of these limits, highlight opportunities in quantum AI and analogue computing, and lay down a roadmap for the adoption of 'Big AI': a synthesis of theory-based rigour with the flexibility of machine learning.

**Introduction**

Artificial intelligence has been hailed as the defining innovation of the 21st century. The current wave of AI hyperbole is traced back to 2012, when Geoffrey Hinton's group used deep convolutional neural networks to win the ImageNet competition - an annual computer vision competition - by a large margin[1]. AI has achieved genuine breakthroughs in protein structure prediction, game playing, and image recognition. Current expectations encompass every endeavour: games, drug discovery, energy management, climate modelling, defence, and even the conduct of science itself[2]. Venture capitalists, policymakers, and researchers alike have spoken of a revolution akin to that spurred by earlier disruptive technologies, from the wheel and railways to electrification and the internet.

Yet there is a yawning gap between aspiration and realization. The MIT Sloan Center for Information Systems Research found that while enterprise AI adoption is widespread, many organisations remain in early maturity stages[3]. Most companies use AI for marketing and customer engagement, rather than for solving core research or engineering challenges.

At the same time, AI was given the ultimate scientific recognition in 2024. The Physics Nobel Prize, shared by Hinton, cited foundational discoveries and inventions that enable machine learning with artificial neural networks,[4] while the chemistry Nobel prize acknowledged AI's role in protein structure prediction and computational protein design[5], suggesting that AI can make a valuable contribution to scientific progress.

That remains to be fully established. The key question is the following: are AI predictions as trustworthy as they are plausible? There is still a gap between expectation and reality, as shown by AI-driven drug discovery startups, which once promised to compress timelines by years. While some stages of drug discovery benefit from machine learning, many do not. AI has helped to shave several months – up to a year in some cases - from the discovery process, which is promising but falls far short of the hyperbole. Although AI models are used for target identification, molecule screening, and retrospective analysis, no AI-discovered molecule has yet received full regulatory approval. There are many claims that, though tantalising, are not yet substantiated in the clinic, such as the discovery of a potential cancer therapy pathway[6]. The demise of BenevolentAI is another example of this cognitive dissonance: once valued at over $7 billion, it collapsed in 2023 amid unmet claims[7]. The saga exposed over-optimism in presuming ML alone could overcome biological, regulatory, and experimental complexity.

Even when it comes to AlphaFold, the Nobel prize winning AI tool, its protein structure predictions accelerate efforts to design drugs but are no substitute for experiments[8]. Moreover, a new generation of 'co-folding' AI models, AlphaFold3, Boltz-1, Chai-1 and RoseTTAFold All-Atom, though impressive, sometimes defy the laws of physics and chemistry. A recent paper warned against uncritical faith in AI and argued for models grounded not just in data but in the principles that govern the natural world.[9] Similarly, when Google DeepMind trumpeted its discovery of 2.2 million new crystalline materials using AI, it seemed to promise a revolution in materials science; instead, critics found some of its digital crystals fantastical or unworkable: machine learning still needs the help of human chemists.[10]

While critiques of AI limitations exist across disciplines - from philosophy to computer science - this perspective offers a novel synthesis. We connect computational physics constraints (floating-point limitations, chaos sensitivity) with epistemological concerns (spurious correlations, lack of world models) and practical failures (drug discovery, materials science). Our central thesis that physics provides the constraints, interpretability, and rigour that AI fundamentally lacks extends to proposing 'Big AI' - a synthesis of physics-informed modelling and machine learning - as an organising framework. We go beyond earlier calls for "physics-informed neural networks" to advocate a broader agenda that includes quantum and analogue computation.

We first review the current state of AI in science, acknowledging both genuine achievements and overstated claims. We then systematically examine fundamental limitations of current AI architectures. Finally, we propose 'Big AI' and outline concrete pathways toward more reliable, interpretable artificial intelligence. We acknowledge that AI has achieved notable successes in specific domains but argue that broader impact requires deeper integration with physical principles.

**Illusions of Intelligence**

The current wave of hyperbole surrounds generative AI. Large Language Models, LLMs (machine learning model trained on massive text datasets to generate and understand human-like language, along with coding and some mathematics), Large Reasoning Models, LRMs (optimized not just for language but for structured reasoning, logic, and problem-solving) and Agentic AI (which can plan, take actions, and pursue goals autonomously). They are often assumed to represent transformative breakthroughs such that (leaving aside the point that there is no widely accepted definition of what we mean by natural intelligence[11]) they are even promoted as on a trajectory toward artificial general intelligence, AGI, a hypothetical

AI with human-level or greater ability to learn, reason, and perform any cognitive task.

However, all these generational AI systems remain rooted in Google's transformer paradigm, a neural network architecture that processes data by paying selective "attention" to various parts of the input at once. This approach has limitations, which we will explore[12].

Rather than only ask what the current generation of AI can do for physics, this article argues that it would be more fruitful to consider what physics can do to improve AI. While some have claimed[13] that big data could lead to the "end of theory," with pattern detection replacing hypothesis-driven science, a more common view is that big data analysis needs a robust theoretical framework to interpret patterns, test hypotheses, and derive meaningful, actionable knowledge. Insights from physics provide grounding, constraints, and interpretability to improve the current generation of AI[14].

**Fundamental Limitations of Current AI Architectures**

Current AI systems suffer from interconnected deficiencies that stem from their training methodology and mathematical foundations. These limitations fall into three broad categories:

*1. Distributional Failures: AI* systems are usually trained and validated on a closed universe of data. But apply them to another universe of data and, though their predictions may be plausible, one cannot be sure that they are correct. There have been high profile instances of AI image recognition failing to work accurately for people from minority ethnic groups, for example, because training sets are dominated by lighter-skinned faces.[15] In healthcare systems, significant racial bias has been found for similar reasons.[16] When AI encounters data outside its training distribution - so-called out-of-distribution inputs - its reliability collapses. The problems with 'out of distribution' AI, which is extrapolating rather than interpolating, could be addressed if uncertainty qualification were easy with AI. But, as we will see, it is not.

Nor are real-world data uniform: they can contain gaps or be skewed or subject to sampling errors. Many machine learning algorithms assume data or noise is normally distributed (Gaussian, or bell-curve shaped) because this simplifies the mathematics, makes models analytically tractable, and underpins classical results (for example, linear regression assumes Gaussian noise). But real-world data is rarely Gaussian. Human language, images, financial markets, biological data —all tend to be heavy-tailed, multimodal, or skewed. If models implicitly or explicitly assume a Gaussian distribution, they can underestimate rare events, fail on marginal cases, or misjudge uncertainty. Modern deep learning often avoids explicit distributional assumptions, yet it still leans on Gaussian scaffolding. In models such as variational autoencoders, for instance, latent variables are explicitly assumed to follow a Gaussian distribution, allowing smooth interpolation in the hidden space. Diffusion models, which now dominate AI image generation, rely on Gaussian noise to corrupt and then reconstruct data. Even in simple regression losses, the common use of mean-squared error implicitly assumes Gaussian noise. A Gaussian (normal) distribution has an undeniable role, but it is far from universal. Indeed, it fails to describe most phenomena where complexity holds sway. The real world is nonlinear and sharp discontinuities can occur.

ML algorithms also assume smooth (differentiable) relationships between the quantities they handle as a matter of convenience, because this allows the use of linear algebra, standard software libraries, and substantial speed-up by GPU accelerators. But the very non-linear systems that they are designed to predict are often riddled with discontinuities. These "jumps" challenge the mathematical assumptions the algorithms rely on, leading to unstable or misleading outputs. As a result, even models that look impressive on benchmark problems may stumble badly when faced with real-world systems that do not behave smoothly.

Humans play a surprisingly important role in how AIs are set up and used, notably in establishing the categories used for classification. That choice is frequently made without any attempt to understand the structural characteristics that underlie the system of interest, with the result that the 'AI system' produced strongly reflects the limitations or biases (be they implicit or explicit, as was the case with classifications used by ImageNet ) of its developers.

*2. Lack of Physical Understanding:* A foundation model is a generative AI system trained on large datasets that can then be adapted and fine-tuned for specific tasks. Fundamentally, however, they do not 'understand' the world as humans do. Astronomers like Johannes Kepler noticed patterns in the night sky that could be used to pinpoint the future locations of planets, and Isaac Newton would later generalise these insights to develop Newtonian mechanics. But AI foundation models cannot yet make the transition from good predictions to meaningful world models[17]. One recent study found, for example, that foundation models trained on the orbital trajectories of celestial bodies consistently fail to apply Newtonian mechanics when adapted to new physics tasks[18]. Though carefully constructed to see if, based on data alone, a foundation model could build a "world model;" it failed to do so. The foundation model had no conception of Newton's law of gravity, even though it could have discovered this law by taking the second derivative of the trajectory data. Instead of generalising laws, the model learned "task-specific heuristics", or shortcuts, producing in the authors' words "nonsense". Akin to Ptolemy's celestial epicycles - circles upon circles that once mapped the heavens with dazzling but misleading precision - LLMs appear 'intelligent' through sheer accumulation of parameters rather than understanding[19]. Big data LLM-based AI is not enough: theory remains indispensable if explanations are required[20].

As a corollary, it is no surprise that, unlike physics-based models, traditional AI does not provide insights and mechanistic understanding—only predictions based on statistical inference. They rely on big data and complex algorithms to identify patterns, but their workings are opaque, even to the system developers. This makes them 'black boxes' that are hard to trust in high-stakes applications like medicine. An LLM may "hallucinate" its way through a series of fluent, sometimes convincing, but factually incorrect steps.[21]

One way forward is the development of forms of AI that are inherently interpretable, meaning the complexity or design of the system is limited so that developers obtain more insights into how it works. Other approaches test how an AI 'black box' works, for example by rerunning an initial model with some inputs changed to work out which ones are most salient.

Since the advent of Large Reasoning Models, LRMs, such as OpenAI's o1 and DeepSeek-R1 and the introduction of reinforcement learning, LRMs are attempting to provide correct explanations. However, they too do not currently provide trustworthy results because of the absence of defined metrics to assess the reliability of such explanations when applied to situations which are less quantitatively assessable than coding and mathematics.

One comparative study of six large language models—DeepSeek, ChatGPT, and Claude, including their reasoning-optimized variants—shows that models tuned for reasoning consistently outperform non-reasoning counterparts in scientific computing and machine learning tasks. However, even these advanced models are prone to ambiguous or incorrect outputs, underscoring the need for improvements in LLMs for scientific problem solving.[22] Another study, The Illusion of Thinking, concluded that LRMs only outperform LLMs in medium complexity tasks and both collapse in accuracy when tackling problems beyond certain complexities, ultimately raising questions about the true reasoning capabilities of LRMs[23]. Although these findings were contested by a paper entitled The Illusion of The Illusion of Thinking, which argued there were shortcomings in experimental design[24], another study (entitled: The Illusion of The Illusion of The Illusion of Thinking) with a better test design concluded 'genuine challenges remain' in the capabilities of LRMs[25].

*3 Digital Pathologies, Spurious Correlations and Degenerative AI*: Typically, digital computers handle four billion rational numbers that range from plus to minus infinity, known as the 'single-precision IEEE floating-point numbers (the 32-bit or FP32 numbers). Yet a significant part of the richness of the real world is only captured by irrational numbers which cannot be represented on any digital computer. In one study, which compared the known mathematical reality of the generalised Bernoulli map to what FP32 computers predict, the results are wrong in many circumstances and catastrophically so in some instances. As a result of the discreteness of floating-point numbers, significant errors can arise in digital computers and the full extent of these errors is not understood.[26] This suggests that renewed focus on analogue computers will be necessary in the long term, and not just because of the soaring power demands of high-performance digital computers.[27] It is hardly surprising that, if we are to achieve AGI, we will need to pay closer attention to how the human brain actually works, not least its analogue nature.

Meanwhile, the current generation of machine learning algorithms typically require hundreds of thousands to trillions or more parameters, which are the connection weights between pairs of "neurons" in a neural network. At the same time, companies such as NVIDIA are using hardware and software that prioritise half-precision and lower precisions to cut memory usage and boost computational speed, handling single and double precision using software emulators. As a result, a reduction in computational performance of high-fidelity simulations is inevitable. However, even if an AI operates in single precision, it depends on more fitting parameters than there are available numbers for its computations.

With more parameters than representable values used in AI, many collapse to the same numerical value due to rounding. This leads to redundancy, where different parameters behave identically, reducing the model's ability to learn nuanced patterns. One mitigation strategy is to use stochastic rounding: instead of always rounding a number up or down to the nearest whole number (or decimal place), that decision is made randomly, based on how close the number is to each option.[28]

Nor do these vast numbers of parameters have intrinsic meaning in the real world. They simply fit inputs to outputs to provide a good match to real-world behaviour. The astronomical number of parameters explains why ML can successfully fit so many arbitrary relationships, like a glorified 'look up' table. But this also accounts for their unreliability, when making predictions based on unseen data.

A central theoretical hurdle facing the use of big data by AI was identified by Cristian S. Calude and Giuseppe Longo in *"The Deluge of Spurious Correlations in Big Data."* As datasets grow, spurious (random) correlations vastly outnumber meaningful ones. LLMs cannot distinguish these, so adding more data reduces the signal-to-noise ratio and contributes to error pileup. Such correlations arise even in randomly generated data: the size of a dataset alone ensures that that the deluge emerges.

Spurious correlations outnumber meaningful ones in very large, high-dimensional data sets. The only way to sort the wheat from the chaff is through the scientific method. This is far from trivial and, particularly in domains where adequate theoretical understanding is lacking, one cannot make this distinction readily.

Yet the companies which have developed the dominant large language models still assert that bigger is better. Partly as a consequence of Calude and Longo's findings, it is plain that making LLMs larger by training them on more data and including vastly more parameters does not guarantee significant improvement[29]. Indeed, the current efforts being undertaken to try to improve LLM performance are purely empirical.

The existing algorithms (based on the transformer architecture) exhibit poor scaling properties. These prevent LLMs from achieving the reliability and accuracy required for scientific applications through brute-force scaling. When using principles from statistical physics, one finds that the incremental improvements to LLM algorithms come at the cost of astronomical compute and energy requirements (tens of gigawatt-hours of electricity now and hundreds soon, underlining why Microsoft struck a deal to restart the Three Mile Island nuclear plant[30]). The nonlinear activation functions within LLMs transform Gaussian inputs into non-Gaussian outputs with fat tails. In this situation, uncertainty in predictions decays much more slowly than would otherwise be expected, further compounding the difficulty of achieving high accuracy. This problem is exacerbated by Calude and Longo's deluge of spurious correlations which increase very rapidly as the size of the data set increases. The outcome can then sometimes be a loop of self-amplifying error - what is sometimes called "degenerative AI" - where predictions degrade with increasing amounts of training data, especially if that data is augmented by data generated previously by AI systems.

A wall confronts large language models[31] and it becomes steeper, and more quickly, when models are trained on low-quality or synthetic data, creating a feedback loop of compounding inaccuracies. To deal with this profound problem, one must understand the behaviour of LLMs based on the theory of non-linear dynamical systems. That is, we need real world physics to explain how they work so we can develop better algorithms than current transformers.

**What Physics Could and Should do for AI**

Physics provides precisely what today's AI lacks: constraints, interpretability, and uncertainty quantification. Combining AI with physics-informed methods often leads to improved performance: Physics-Informed Neural Networks (PINNs) outperform pure data-driven approaches in fluid dynamics[32], extracting quantitative information for which direct measurements may not be possible.[33] In weather forecasting, claims made for AI outperforming conventional methods are contentious. AI does not do as well as physics-based models, particularly when chaos is relevant at short time scales: they cannot simulate the 'butterfly effect'.[34] Nor can AI forecast weather events beyond the scope of existing training data (such as events that are so rare they are so called gray swans), which might exclude unprecedented heat waves, floods or hurricanes.[35] What it is mainly useful for being able to make predictions using "inference" much faster. NowcastNet, blends physics-based forecasting, based on fluid flow equations, with deep learning to provide 'nowcasting' of precipitation with a higher success than traditional numerical models.[36] [37]

Chaos is also an issue when it comes to molecular dynamics, MD, which has been unappreciated by the MD community: nonlinearity undermines the ability to train an AI system on one off sets of data. AI predictions made fail even when interpolating (as opposed to extrapolating) because any real molecular system (say the space of all molecular structures), is vastly more complicated than the AI can have any knowledge of. Even here, introducing physics-based methods to ensure predictions of AI are scientifically reasonable can create a virtuous circle.[38]

AI systems are only as robust as the assumptions on which they rest and the data they ingest. Poorly curated, biased, or synthetic datasets can introduce systemic errors, especially when used recursively in model training. Rigorous data curation, provenance tracking, and annotation standards are essential to prevent feedback loops amplifying misinformation.

AI requires rigorous verification, validation, and uncertainty quantification (VVUQ), the triad that underpins trustworthy modelling in applied mathematics, statistics, and engineering. While VVUQ is well-established for traditional engineering simulations, it is a subject of active research, for example to deal with large uncertain parameter spaces (for possibly hundreds to trillions of parameters). In AI and machine learning, this field is still in its infancy, hampered by models whose workings are opaque, sprawling in dimension, and detached from physical reality[39].

For a physics-based model, in contrast, VVUQ reveals precisely how to improve predictions. Unlike physics-based models, however, machine learning algorithms lack a direct mapping between model parameters and physical quantities, making interpretability and refinement difficult. Even if we can pinpoint connection weights which are responsible for most of the uncertainty in a machine learning algorithm, there is no way of knowing how or what to adjust to reduce the uncertainty. The principal way claimed to improve predictions from AI systems rests on a simple idea: "give me more data", but we have already explained why that approach is neither necessary nor sufficient.

AI researchers sometimes resort to *ad hoc* techniques such as a so-called Monte Carlo dropout procedure, which estimates predictive uncertainty by systematically omitting an increasing fraction of the neurons within each hidden layer during inference, and Bayesian neural networks, which treat weights as probability distributions rather than fixed numbers. However, these methods lack precision, do not comply with the tenets of systematic UQ, and typically underestimate the true uncertainty in these AI systems. In recent work underway with Wouter Edeling, one of us (PVC) has found that a deep active subspace approach taken from state-of-the-art UQ methodology, building on related preliminary work, may provide a means of dramatically reducing the number of parameters required to capture the key behaviour of AI systems.[40]

The prize of this effort is clear. If and only if AI-based predictions pass muster, in terms of VVUQ, they will become "actionable" – you can use them to make decisions in critical circumstances, for instance when caring for patients, because you can be confident that their predictions are indeed reliable (within a specified uncertainty bound). With reliable uncertainty bounds, AI outputs can also be safely "actionable" in science.

Aside from analogue computation, another way to tame chaos is to blend quantum-inspired machine learning with classical dynamical systems theory[41]. Traditional machine learning models, when applied to chaotic systems, often require vast computational resources and still fall short of capturing long-term dynamics. These systems are sensitive to initial conditions, and small errors can quickly snowball.

The solution lies not in brute force but in elegance, in the form of understanding. Here we are referring to a Quantum Circuit Born Machine (QCBM), a probabilistic model inspired by the Born rule in quantum mechanics[42,43]. QCBMs learn patterns in data using principles from quantum physics. Because they work in the huge mathematical space that quantum systems naturally occupy, Hilbert space, they can capture relationships in data that classical machine-learning models might miss. And they can do this with far fewer adjustable settings. The patterns they learn - quantum priors" - can then be fed into ordinary AI models to help them make better predictions. . For example, QCBMs can be used to learn the high-dimensional energy landscapes that govern molecular dynamics, helping to predict stable molecular conformations or rare transition states more efficiently than classical sampling methods. One application has been to small molecule design of KRAS inhibitors[44].

Quantum computing remains in its infancy, though there are encouraging hints of quantum advantage in this domain. In the meantime, quantum-inspired machine learning (QIML) - including models like QCBMs - offers a useful middle ground: it borrows ideas from quantum physics but still runs on ordinary computers. By embedding physical insight into machine learning architectures, smarter - not simply bigger - models may be the key to unlocking complex phenomena.

However, there are limits. Scaling these methods to the enormous chemical libraries used in drug discovery is still difficult. And until analogue AI becomes mainstream, today's digital AI systems remain poor at dealing with truly chaotic, highly complex systems. Their predictions tend to fall apart the further they look into the future, they struggle to measure uncertainty well, and they sometimes mistake short-term patterns for real underlying behaviour. These shortcomings mean AI is better seen as a tool for probing chaos locally, rather than conquering it outright.

13This work is part of a broader movement to integrate physics and AI, a trend that promises to reshape scientific computing[45]. With applications being investigated in a wide range of domains, from climate systems to biological processes, the fusion of quantum ideas with machine learning may offer a new path forward—one that is not only computationally efficient but also grounded in the laws of nature.

Physics provides a rigorous theoretical framework for understanding learning algorithms, through the application of non-linear dynamical systems.
theory and statistical mechanics. Physics offers the constraints needed to sift meaningful patterns from spurious ones. Embedding symmetries, conservation laws, invariances and understanding into model design can help prevent AI from over-relying on misleading patterns. For applications requiring high reliability, notably in medicine, AI needs to integrate scientific principles and domain-specific knowledge, whether physical, chemical, or biological. Physics-informed models are not a panacea, however. They require well-understood governing equations and may struggle in domains such as economics and social systems where such laws are unknown or poorly defined.

The ultimate aim should be physics-informed learning — what we call "Big AI"[46] — which blends theory with machine learning. To achieve Big AI ideally means embedding conservation laws, symmetries, and invariances directly into model architectures; combining interpretable physics equations with flexible ML components: certifying uncertainty with VVUQ inherited from physics-based modelling; providing mechanistic interpretability, such that each parameter maps to physical quantities or processes; and theory-guided learning, where domain knowledge constrains parameter spaces and prevent spurious correlations. Ultimately, analogue computation, conventional and quantum, should be the substrate as it should often be the source of the "ground truth" against which AI predictions should be made. The goal is not to replace either physics or ML, but to create a new paradigm wherein the whole is greater than the sum of these two parts.

This provides interpretability and mechanistic insight (physical constraints, laws, conservation); better uncertainty quantification (because physics-based models allow error bounds, sensitivity to input variation, and so on); improved generalization, especially outside data-rich settings, since the physics component embeds known invariances or behaviour.

| Feature | Pure AI | Physics-Informed AI |
|---|---|---|
| Interpretability | Low | High |
| Uncertainty Qualification | Weak | Strong |
| Scalability | High | Moderate |
| Mechanistic Insight | None | Embedded |
| Data Dependence | High | Lower |

Caption: *Pure AI learns from data; Big AI learns from nature.*

**A Roadmap for Big AI**

To realise the benefits of physics informed AI, we propose a research agenda across three timescales:

Near-term (1-3 years): Develop new AI and open physics-informed algorithms with improved scaling performance and more reliable error metrics; Standardize VVUQ protocols for ML in scientific computing; Establish community-defined benchmarks comparing physics-informed with pure ML predictions across multiple domains; Understand in more quantitative terms the impact of the deluge of spurious correlations to determine efficient means of dealing with it.

Medium-term (3-7 years): Scale quantum-inspired machine-learning algorithms to industrially relevant problems; Develop hybrid analogue computing architectures or practical use; Integrate causal inference with physics constraints for improved generalization and reliable reasoning.

Long-term (7-15 years): Achieve fault-tolerant quantum computing for scientific ML; Develop high accuracy, high precision analogue systems for general purpose computing; Establish an AGI theory grounded in physical law, not in the more speculative promise—championed by AI maximalists—that an algorithm might one day infer or even invent the universe's rules from pure data.

**Why Physics Must Shape the Future of AI**

The next revolution in AI will not be driven by scale alone, but by its consistency with and understanding of the laws of nature. AI has contributed to genuine scientific advances, but its current incarnation often provides glib answers that raise troubling questions about mechanistic insight, scalability and reliability. As a result, claims of the broader promise of current AI as a step towards AGI remain unconvincing. The subjective elements of modern AI, its reliance on vast numbers of meaningless parameters, black-box models, and very limited uncertainty quantification have led to many failures, especially in extrapolation beyond its own training data but also not infrequently when interpolating. The laws of nature and rigorous theory are essential correctives. Embedding physical laws, using hybrid models, constraining parameters, improving uncertainty estimation, and exploring quantum and analogue computational paradigms are pathways to a more mature, dependable, and scientifically credible AI.

To move beyond pattern recognition and toward genuine understanding, AI must be grounded in scientific laws, theory, and curated data. This demands a new kind of collaboration - between theoreticians, computer scientists, and domain experts - to co-design models that are not just powerful, but principled, interpretable, and physically constrained. Physics-informed AI – Big AI - also offers a path to more auditable and trustworthy systems, especially as AI increasingly influences decisions in science, the workplace and medicine. To safeguard these advances, we must also build robust governance frameworks that ensure transparency, accountability, and safety. The future of AI will be defined not only by advances in algorithms, but by its integration with the fundamental laws of nature and the discipline of the scientific method.

---

**Acknowledgements**

We thank Sauro Succi and a reviewer for helpful comments. The authors acknowledge funding from (i) UKRI-EPSRC for the UK High-End Computing Consortium (EP/R029598/1), the Software Environment for Actionable & VVUQ-evaluated Exascale Applications (SEAVEA) grant (EP/W007762/1), the UK Consortium on Mesoscale Engineering Sciences (UKCOMES grant no.

EP/L00030X/1), and the Computational Biomedicine at the Exascale (CompBioMedX) grant (EP/X019276/1); (ii) the European Commission for EU H2020 CompBioMed2 Center of Excellence (grant no. 823712). RH is a member of the UKRI-Medical Research Council.

**Affiliations**

Peter Coveney

Centre for Computational Science, Department of Chemistry, University College London, London WC1H 0AJ, U.K.

Advanced Research Computing Centre, University College London, London WC1H 0AJ, U.K.

Institute for Informatics, Faculty of Science, University of Amsterdam, 1098XH Amsterdam, The Netherlands

*Corresponding to p.v.coveney@ucl.ac.uk

Roger Highfield

Science Museum, Exhibition Road, London SW7 2DD, U.K.

Sir William Dunn School of Pathology, University of Oxford, Oxford OX1 3RE, U.K.

Department of Chemistry, University College London, London WC1h 0AJ, U.K.

Member, Medical Research Council